\documentclass[10pt]{article}
\usepackage[top=0.85in,marginparwidth=2in]{geometry}

\usepackage{amsmath,amsfonts}
\usepackage{algorithmic}
\usepackage{algorithm}
\floatstyle{plaintop} \restylefloat{algorithm}
\usepackage{array}
\usepackage{textcomp}
\usepackage{stfloats}
\usepackage{color}
\usepackage{url}
\usepackage{verbatim}
\usepackage{listings}
\usepackage{booktabs}

\usepackage{bm}             
\usepackage{mathrsfs}       

\usepackage{cancel}         

\usepackage[utf8]{inputenc}

\usepackage{cite}

\usepackage{nameref,hyperref}

\makeatletter
\renewcommand{\@biblabel}[1]{\quad#1.}
\makeatother

\usepackage{lastpage,fancyhdr,graphicx}
\usepackage{epstopdf}

\usepackage{color}

\usepackage{graphicx}

\usepackage{sidecap}
\usepackage{booktabs}

\begin{document}
\vspace*{0.35in}

\begin{flushleft}
{\Large
\textbf\newline{Neuromorphic on-chip reservoir computing with spiking neural network architectures}
}
\newline
\\
Samip Karki\textsuperscript{1},
Diego Chavez Arana\textsuperscript{2},
Andrew Sornborger\textsuperscript{3},
Francesco Caravelli\textsuperscript{4}
\\
\bigskip
\bf{1} Center for Integrated Nanotechnologies, Los Alamos National Laboratory, Los Alamos 87544 NM, USA
\\
\bf{2} Department of Electrical and Computer Engineering
 New Mexico State University, Las Cruces, NM 88003, USA\\
\bf{3} Information Sciences (CCS-3), Los Alamos National Laboratory, Los Alamos 87544 NM, USA
\\
\bf{4} Theoretical Division (T-4), Los Alamos National Laboratory, Los Alamos 87544 NM, USA
\\
\bigskip
* caravelli@lanl.gov

\end{flushleft}

\section*{Abstract}
Reservoir computing is a promising approach for harnessing the computational power of recurrent neural networks while dramatically simplifying training. This paper investigates the application of integrate-and-fire neurons within reservoir computing frameworks for two distinct tasks: capturing chaotic dynamics of the H\'enon map and forecasting the Mackey-Glass time series. Integrate-and-fire neurons can be implemented in low-power neuromorphic architectures such as Intel Loihi. We explore the impact of network topologies created through random interactions on the reservoir's performance. Our study reveals task-specific variations in network effectiveness, highlighting the importance of tailored architectures for distinct computational tasks. To identify optimal network configurations, we employ a meta-learning approach combined with simulated annealing. This method efficiently explores the space of possible network structures, identifying architectures that excel in different scenarios. The resulting networks demonstrate a range of behaviors, showcasing how inherent architectural features influence task-specific capabilities.
We study the reservoir computing performance using a custom integrate-and-fire code, Intel's Lava neuromorphic computing software framework, and via an on-chip implementation in Loihi. We conclude with an analysis of the energy performance of the Loihi architecture.


\section*{Introduction}

Neuromorphic computing is an innovative and promising approach to computational systems that draws inspiration from the structure and function of the human brain. Although the word ``neuromorphic" was originally intended for CMOS circuits that mimic the brain's behavior \cite{mead1989analog,salam}, the term is slowly shifting towards a broader umbrella term that incorporates a variety of different approaches, including alternative optimization techniques that are biologically inspired \cite{Ielmini2018,Sebastian2020,reviewCarCar,johannson2012slime}.

Mimicking the neural connections and synapses found in the brain, neuromorphic computing aims to create highly efficient and powerful computing architectures, with the object in mind to tackle computational problems suited for analog devices. The list of open and proprietary neuromorphic chips available to use for research purposes is growing steadily \cite{Greengard2020,Yamane2023} due to the possibility of applying machine learning algorithms directly on a low-power device \cite{Painkras2012,Sengupta2019,Billaudelle2021}.


Conventionally, backpropagation (BP), along with other learning algorithms, has remained indispensable for facilitating supervised learning in artificial neural networks (ANNs). While the question of whether BP operates within the brain remains an open question, the BP approach effectively addresses the challenge of connecting a global objective function to localized synaptic adjustments within a network.
Deep learning, a central tool in modern machine learning, hinges on layered, feedforward networks reminiscent of the early strata of the visual cortex. These networks encompass threshold nonlinearities at each layer, mirroring mean-field approximations of neuronal integrate-and-fire models \cite{indiveri}. 

More recently, there has been a rapid surge of interest in neuromorphic computing, which has emerged from the re-imagining of classical algorithms geared toward learning, optimization, and control, utilizing event-based or spike-based information processors. The latter in particular is assuming an increasingly important role within the domains of both theoretical neuroscience \cite{Izhikevich2007} and neuromorphic computing (NC). In the case of NC, this is because there are now available chips able to perform a low-power emulation of interconnected leaky integrate-and-fire neurons \cite{Gerstner2014}. For instance, a fundamental aspect of brain learning involves the modulation of synaptic strengths between neurons and neural populations.  This necessity holds true even within the confines of individual neural circuits, whether biological or artificial.  In particular, this trend has been at the core of recent versatile computing architectures, exemplified by Intel's Loihi (and Loihi2) neuromorphic processor \cite{davies2018loihi}. Although the translation of feedforward networks into neuromorphic hardware poses reasonable challenges, the considerably more computationally intensive task of on-chip training has proven to be elusive, mainly due to the intricate nature of the backpropagation algorithm, despite recent impressive progress \cite{sornborger,Shen2022}.

For this reason, we seek an alternative to feed-fordward architecture, e.g. to consider Recurrent Neural Networks (RNNs) \cite{haykin}.  
Training an RNN involves initializing parameters like weights and biases: Backpropagation Through Time (BPTT) is performed by calculating gradients of loss for parameters over multiple time steps. Typically, weights are updated using optimization algorithms like gradient descent with the addition of stochasticity in the dynamics. Training traditional RNN is still time-intensive. From an energy consumption perspective, training a neural network is time- and energy- consuming. For this reason, the combination of simpler machine learning paradigms with energy-efficient architectures can be important in settings where an energy budget can be of importance. Reservoir Computing (RC) \cite{Jaeger2004} is a powerful concept within the realm of RNN. Unlike conventional RNNs where all connections are learned through training, Reservoir Computing maintains a fixed and randomly initialized network of interconnected nodes, referred to as the ``reservoir" (which acts more generically as a dynamical system inducing some form of nonlinearity at the output).
The key advantage of Reservoir Computing lies in its simplicity and efficiency. As the reservoir connections remain fixed, the training process is solely focused on optimizing the output weights, which is a linear regression, drastically reducing the complexity of learning at the cost of performance. This results in faster training times and the ability to handle larger datasets without encountering issues like vanishing gradients. Additionally, if the dynamics of the reservoir is sufficiently rich, it will still enable to capture the intricate temporal dependencies in data, making it particularly well-suited for time series prediction, speech recognition, and various sequential tasks. A simple way to understand the workings of a reservoir is the automatic generation (through the input signal) of a temporal basis over which the regression can be performed. 

In the present paper, we study the implementation of a reservoir computer in spiking neural networks of the leaky integrate-and-fire type. 
One of the initial proposals for the functioning of a reservoir computer was the echo state (or fading property in the more general setting \cite{sheldoncar}), e.g. the property of the reservoir to eventually forget its initial state or configuration.
However, although it was initially thought that as long as a reservoir satisfies the echo state or fading property a random architecture would be sufficient\cite{ieeespikres}, this property has become of secondary importance today. 
In fact, by optimizing the network's size and connectivity, a reservoir can become better suited to perform specific tasks. Below, we show that it is important to select the architecture properly (by evaluating the error on a prediction task), and also to allow generalization. 
Reservoir computers with biologically plausible models (such as integrate-and-fire) are sometimes known as Liquid State Machines \cite{maass2002realtime,deckers2022extended}. However, it is rarely the case that the interaction networks are optimized to fit the task at hand \cite{patino-saucedo2022liquid,tian2021neural}, and are indeed often assumed to be random \cite{patel2022liquid}. Although there have been implementations of reservoir computing models with Loihi \cite{Gaurav2022,Gaurav2023}, this paper is the first that optimizes the architecture. In the case of reservoir computers, optimization of the network is often important to improve the performance at a certain task \cite{sheldoncar}, alongside tuning the system to be near the edge of chaos or some form of instability \cite{ivanov2021increasing,Carroll2020,zhultl2}.

The optimization of the network architecture is performed both by careful consideration of what type of architecture is necessary to process information (sparse vs dense), and via an automatized method using meta-learning, (also known as ``learning to learn" \cite{metal}).  Meta-learning has been already applied in reservoir computing with nanowires with discrete success \cite{zhultl, zhultl2}.
Here, the architecture of the reservoir network plays a critical role in its performance. As we show in the present manuscript, meta-learning can be harnessed to automate the process of optimizing the reservoir architecture, thus streamlining the task of designing effective reservoir networks. In particular, as neuromorphic chips such as Loihi \cite{davies2018loihi}, SpiNNaker \cite{Painkras2012,rhodes2018spynnaker} and TrueNorth \cite{osti_1405258} become available, and energy efficiency becomes important, adapting the architecture to the task can be important for energy applications.

In the present paper, we study a model of reservoir computer that makes use of the nonlinearity of the integrate-and-fire model, and thus can be directly implemented on a Loihi architecture \cite{intel-labs-lava-2023, davies2018loihi}. The goal of this paper is to test case various architectures on a time prediction performance, optimizing the integrate-and-fire connections between the neurons first by hand, using random network models and meta-learning.

\section*{The model}
\subsection*{Integrate and fire neurons}
 The integrate-and-fire neuron is a simplified model used to describe the behavior of a spiking neuron. It integrates incoming currents over time and generates a spike when the membrane potential reaches a certain threshold.

The membrane potential of the integrate-and-fire neuron can be described by the following differential equation:

\begin{equation}
    \tau_m \frac{dV}{dt} = - (V - V_{\text{rest}}) + RI(t) \; ,
\end{equation}
where  $V$  is the membrane potential of the neuron, $\tau_m$  is the membrane time constant, $V_{\text{rest}}$  is the resting membrane potential,  $R$  is the membrane resistance, while  $I(t)$  is the input current at time  $t$.
When the membrane potential reaches the threshold value, $V_{\text{thresh}}$, the neuron generates a spike and the membrane potential is reset to $V_{\text{reset}}$.

Integrate-and-fire neurons can be interconnected to form networks. The interactions between neurons are typically represented using a synaptic weight matrix. We consider a network of $N$ integrate-and-fire neurons. The dynamics of the membrane potential $V_i$ of neuron $i$ can be described by:

\begin{equation}
    \tau_m \frac{dV_i}{dt} = - (V_i - V_{\text{rest}}) + \sum_{j=1}^{N} w_{ij} S_j(t) \; ,
\end{equation}
where $w_{ij}$  is the synaptic weight from the $j$'th to the $i$'th neuron, and  $S_j(t)$  is the value of the spike train of neuron $j$  at time  $t$. The spike train $S_j(t)$ is a series of delta functions representing the spike times of neuron $j$. When the membrane potential $V_i$ of neuron $i$ reaches the threshold value, $V_{\text{thresh}}$, the neuron generates a spike, and $V_i$ is reset to $V_{\text{reset}}$.

Loihi's neuromorphic architecture is particularly well-suited for implementing SNNs because it implements the spiking neuron model directly in hardware. In traditional neural network architectures, like feedforward or convolutional neural networks, neurons typically use continuous activations, and computations are performed using floating-point operations. However, spiking neural networks, which are more biologically inspired, use discrete spikes for communication and computation. This allows for event-driven processing, which can lead to significant energy efficiency gains compared to traditional architectures.

Loihi's hardware additionally allows for efficient simulation and execution of SNNs. Each Loihi chip contains a large number of cores, and each core can simulate multiple spiking neurons. The chip's architecture takes advantage of parallelism and event-driven processing to perform computations more closely resembling the brain's functioning. This makes Loihi well-suited for running real-time, low-power, and highly scalable SNNs.

The integrate-and-fire neurons implemented on Loihi can be interconnected to form complex spiking neural networks, enabling a wide range of neuromorphic applications, including pattern recognition, sensor processing, and time-series analysis, among others.

\subsection*{The H\'enon Map and the Mackey-Glass time series} 

\textbf{The H\'enon map}. The H\'enon map is a two-dimensional discrete-time dynamical system that exhibits chaotic behavior. It is defined by the following equations:
\begin{align*}
    x_{n+1} &= y_n + 1 - a x_n^2 \\
    y_{n+1} &= b x_n
\end{align*}
where $x_n$ and $y_n$ represent the state of the system at time step $n$, and $a$ and $b$ are parameters of the map. 

If one substitutes the equation for $y_n$ into the equation for $x_{n+1}$ one obtains the  following one dimensional time-series:

\begin{align*}
    x_{n+1} &= 1 +b x_{n-1} - a x_n^2
\end{align*}

This one-dimensional time-series was used as a benchmark for our explorations. Later in the paper, the specific task we explored with the H\'enon Map dataset was to predict $x_{n+1}$ with information of only $x{n}$.

The H\'enon map is a simple yet powerful example of a chaotic system. Depending on the values of $a$ and $b$, it can display regular, periodic, or chaotic behavior.

\begin{figure}
    \centering
    \includegraphics[scale=0.2]{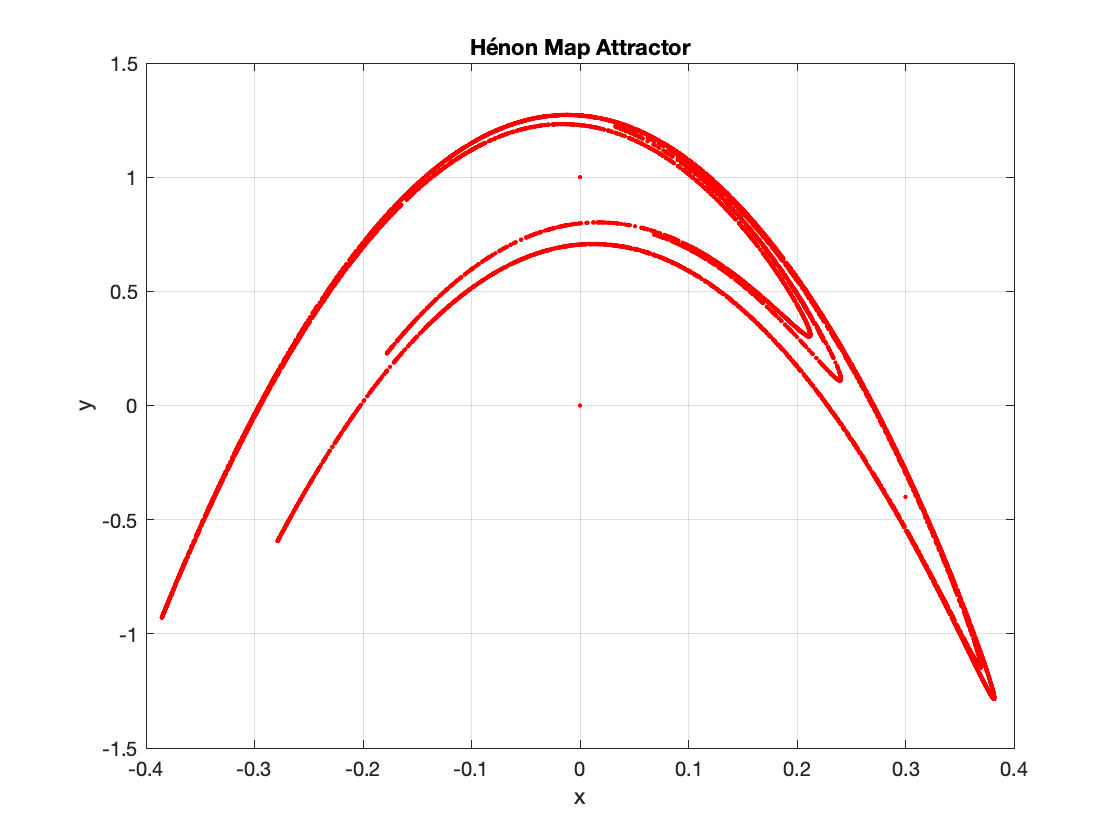}
    \caption{Attractor of the H\'enon map, for $a = 1.4$,
$b = 0.3$, starting from $x=y=0$.}
    \label{fig:henon}
\end{figure}

The transition to chaos in the H\'enon map is achieved by varying the parameter $a$ while keeping $b$ constant. For certain values of $a$ and $b$, the H\'enon map may have fixed points where the state of the system remains constant over time. These fixed points occur when $(x_n, y_n) = (x_{n+1}, y_{n+1})$, indicating an equilibrium state. In this case, the system exhibits regular behavior.
As $a$ is increased beyond a certain critical value, the H\'enon map can exhibit periodic behavior characterized by limit cycles. Limit cycles are closed trajectories in the state space to which the system converges, and the state of the system repeats after a certain number of iterations. The period of the limit cycle increases with higher values of $a$.

\textbf{The Mackey-Glass oscillator}. At a specific critical value of $a$, the H\'enon map undergoes a period-doubling cascade and enters a regime of chaotic behavior. In this chaotic region, the system's trajectories become highly sensitive to initial conditions. Tiny differences in the initial state can lead to vastly different trajectories over time. This sensitivity to initial conditions is a hallmark of chaos.
In the chaotic region, the system's state appears random and unpredictable, yet the dynamics are deterministic. The H\'enon map's chaotic behavior is characterized by a strange attractor, which is a non-periodic, self-repeating trajectory in the state space. The strange attractor has a fractal structure and reveals the underlying complexity of the chaotic system.
The transition to chaos in the H\'enon map demonstrates how small changes in the parameters of a simple mathematical map can lead to a drastic change in the system's behavior, from regular to periodic and finally to chaotic. This sensitivity to parameters is a fundamental property of chaotic systems and has significant implications in various fields, including physics, engineering, and biology.

A second time-series that we explored was the Mackey-Glass Oscillator:

\begin{align*}
\frac{dx(t)}{dt} = \frac{\beta x(t - \tau)}{1 + x(t - \tau)^\eta} - \gamma x(t)
\end{align*}

Depending on the values of the parameters, the Mackey-Glass oscillator can display a wide range of nonlinear dynamical behavior. The value of $\tau$, in particular, controls whether or not the behavior enters the chaotic regime. In our exploration, we considered the oscillator's chaotic behavior with the following parameters: $\beta = 0.2, \gamma = 0.1, \eta = 10$ and $\tau = 18$. The dynamics of the time series are generated with Euler's method and a time step of $0.150$.

A time series $x_n$ is computed from the continuous differential equation by considering the value of $x(t)$ at every time interval $\Delta t = 3$. That is, $x_n = x(t)$ and $x_{n+1} = x(t+\Delta t)$.

\subsection*{Reservoir computing}
Let us now introduce the reservoir computing (RC) framework. 
We denote the input data as a sequence of $T$ vectors $\mathbf{u}_n \in \mathbb{R}^N$ at discrete time steps $n=1,2,\ldots,T$. In its typical implementation, the reservoir is a randomly connected dynamical system with $M$ recurrently connected neurons, and its dynamics can be described as follows:
\begin{equation}
    \mathbf{x}_{n+1} = f(\mathbf{W}_{\text{in}} \mathbf{u}_n + \mathbf{W}_{\text{res}} \mathbf{x}_n)
\end{equation}
where $\mathbf{x}_n \in \mathbb{R}^M$ is the state vector of the reservoir at time step $n$, $f(\cdot)$ is an element-wise activation function, $\mathbf{W}_{\text{in}} \in \mathbb{R}^{M \times N}$ is the input weight matrix, and $\mathbf{W}_{\text{res}} \in \mathbb{R}^{M \times M}$ is the recurrent weight matrix.
The reservoir is described as a network of interacting dynamical components typically constructed as a random graph with sparsely and randomly connected elements. Random connectivity ensures that the reservoir exhibits rich dynamic behavior, making it capable of capturing complex temporal patterns in the input data. The reservoir dynamics create a high-dimensional representation of the input data in $\mathbf{u}_n$. To utilize this representation for a specific task, a linear readout layer is trained to map the reservoir's state to the desired output $\mathbf{y}_n \in \mathbb{R}^P$ at each time step:

\begin{equation}
    \mathbf{y}_n = \mathbf{W}_{\text{out}} \mathbf{x}_n
\end{equation}
where $\mathbf{W}_{\text{out}} \in \mathbb{R}^{P \times M}$ is the output weight matrix, and $P$ is the dimensionality of the output.
 The training of the RC system involves two main steps: reservoir training and readout training.  The output weight matrix $\mathbf{W}_{\text{out}}$ is trained using supervised learning on the reservoir states and the corresponding target output data. This training step tunes the readout layer to generate accurate predictions for the given task. Below, we focus on predicting the $n+1$'st point. For the H\'enon map, this is a rather simple step but given this fact, this also shows why for certain tasks random networks perform more poorly than simpler networks.  For this purpose, our initial attempt was to consider random, Erd\H{o}s-R\'enyi networks of integrate-and-fire neurons. However, it became immediately clear that sparsity was important for this architecture to make good predictions.


In the case of the spiking reservoir network that we explored, the activation function $f(\cdot)$ represents the spikes accumulated during the $n$'th time-step. That is, $x_n$ is a vector with $M + 1$ elements which represents the number of spikes produced by each of the $M$ neurons in the network for a given $\mathbf{u}_n$. In addition, a bias term of $1$ was added as the first element of $x_n$.

During the training process, the reservoir state matrix, $X$, was assembled with each column of $X$ representing $x_n$ for every $n$. The output weight matrix, $\mathbf{W}_{\text{out}}$ was computed with:
\begin{equation}
    \mathbf{W}_{\text{out}}= \mathbf{Y} (\mathbf{X})^+
\end{equation}
where $Y$ is the matrix made of $\textbf{y}_n$ at every $n$ as columns and $^+$ represents the Moore-Penrose pseudo-inverse.
\subsection*{Spatially Encoding Time-Series Data}
Encoding of time-series data was done spatially in the reservoir network. The network of $M$ neurons will have $M_{in} < M$ input neurons. Depending on the value of a given input $u_n$, a single input neuron solely will receive an input current of $I_0$ for $\delta$ discrete time-steps of the time-evolution of the network. To assign values of $u_n$ to the $M_{in}$ input neurons, $u_n$ is discretized in the following way. The values of the time series are scaled such that the minimum of $u_n$ and the maximum of $u_n$ are mapped to $1$ and $M_{in}$, respectively, and rounded to the nearest integer. Thus, a new sequence, $u_n'$ has integer values which correspond to the indexes of the $M_{in}$ input neurons. For example, if $u_n' = 5$, the neuron with the index of $5$ will receive the input current $I_0$ for $\delta$ time-steps. 

\begin{figure}
    \centering
    \includegraphics[scale=0.5]{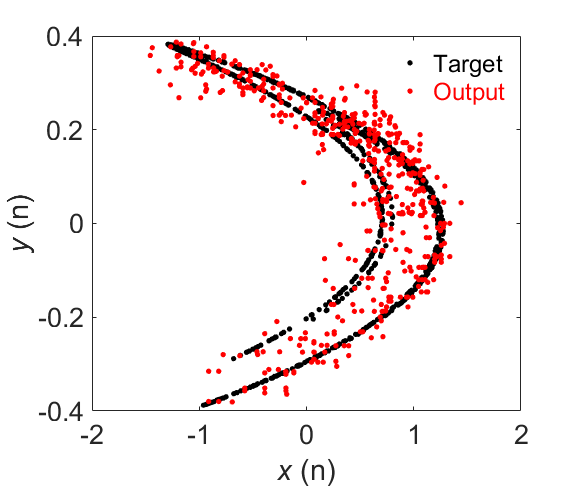}
    \caption{2-D representation of results of H\'enon map task with  Erd\H{o}s-R\'enyi graph network. This simulation included a network of $M=100$ neurons with a connection probability between any two neurons of $2/M$. The input was spatially encoded as described in the main paper. This random graph produced $NRMSE = 0.178$}
    \label{fig:ergraphs}
\end{figure}
\section*{Results}

We now describe the performance of a reservoir computer with a variety of network architectures.

\textbf{Results for  Erd\H{o}s-R\'enyi -} After having tried nearly fully connected interaction graphs, we focused our attention on  Erd\H{o}s-R\'enyi graphs \cite{erdos1959random}. 
The Erdős-Rényi random graph model is a mathematical framework for generating random graphs. Here, we consider the
\textbf{G(n, p) Model}: In this variation, a graph with nodes is generated. Each potential edge between node pairs is included in the graph with an independent probability $p$, resulting in a graph with a random structure. The key characteristics of the Erd\H{o}s-R\'enyi random graph model include the changing average shortest path lengths based on edge density. The model serves as the null model for many random graph experiments with randomly structured networks. 
For $p = 2/M$, where $M = 100$, we can see in Fig. \ref{fig:ergraphs} that the attractor of the H\'enon map can be poorly reconstructed from an Erd\H{o}s-R\'enyi graph.

\begin{figure}
    \centering
    \includegraphics[scale=0.36]{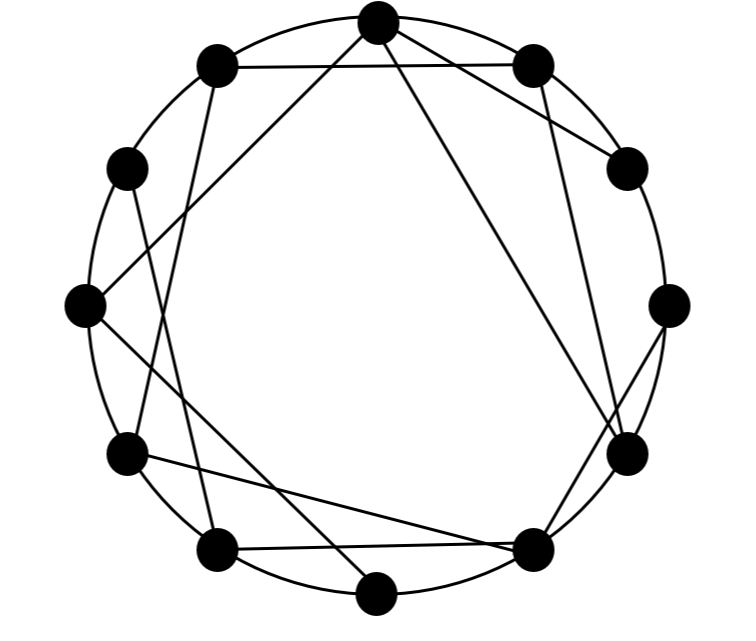}     \includegraphics[scale=0.35]{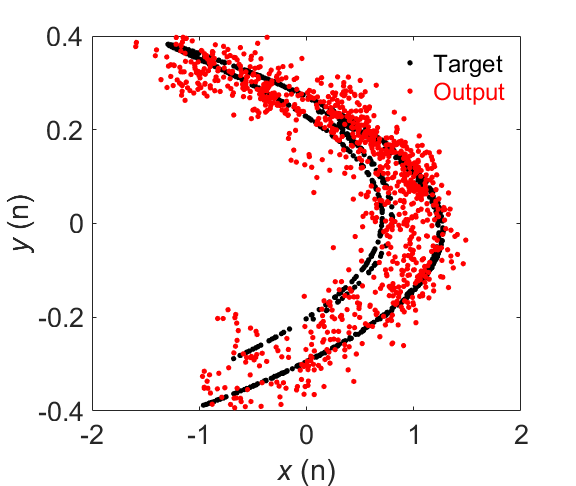}

    \caption{ \textit{Left:} An example of random small world. On the right are the results of the H\'enon map tasks when the network. Every neuron in the small world network was connected to its immediate neighbors and additional connections were added with a probability of $2/M$, where $M = 100$. Input was encoded with $M_{in} = 50$ neurons, these input neurons occurred in every other neuron around the circle. This graph produced a poor $NRMSE$ of $0.204$}
    \label{fig:small-world}
\end{figure}

\textbf{Results for Small-World Networks}  
The internal nodes of the system were generated using an algorithm inspired by small-world networks. We describe first the Watts-Strogatz algorithm for the generation of small-world networks \cite{watts1998collective}. 
One begins with a \textit{regular ring lattice:} one begins with a regular ring lattice, where nodes are arranged in a circle and each node is connected to its nearest neighbors. This initial structure promotes high local clustering but results in longer path lengths for long-range connections. In the second part of the algorithm, a \textit{random rewiring} occurs. This step introduces controlled randomness by adding new connections by rewiring edges at random with probability $p$, while preserving the total number of edges. In this step, we deviated from the Watts-Strogatz network. For each random pair of nodes, we add new connections with a certain probability $p$ on top of the ring, instead of rewiring.  This step introduces shortcuts that reduce path lengths by however preserving the maximum path length on the graph. Thus, both short small world and long ring paths are preserved.  We explored networks with $M = 100$ nodes. Each node was connected to its immediate neighbors with additional connections made with a probability of $2/M$. The input encoding was done with every other neuron in the ring acting as an input neuron. The results are shown in \ref{fig:small-world}. As it can be seen, this network, like the  Erd\H{o}s-R\'enyi network, does a poor job of reconstructing the attractor.

\subsection*{A working network architecture}
\begin{figure*}
    \centering
    \includegraphics[scale=0.17]{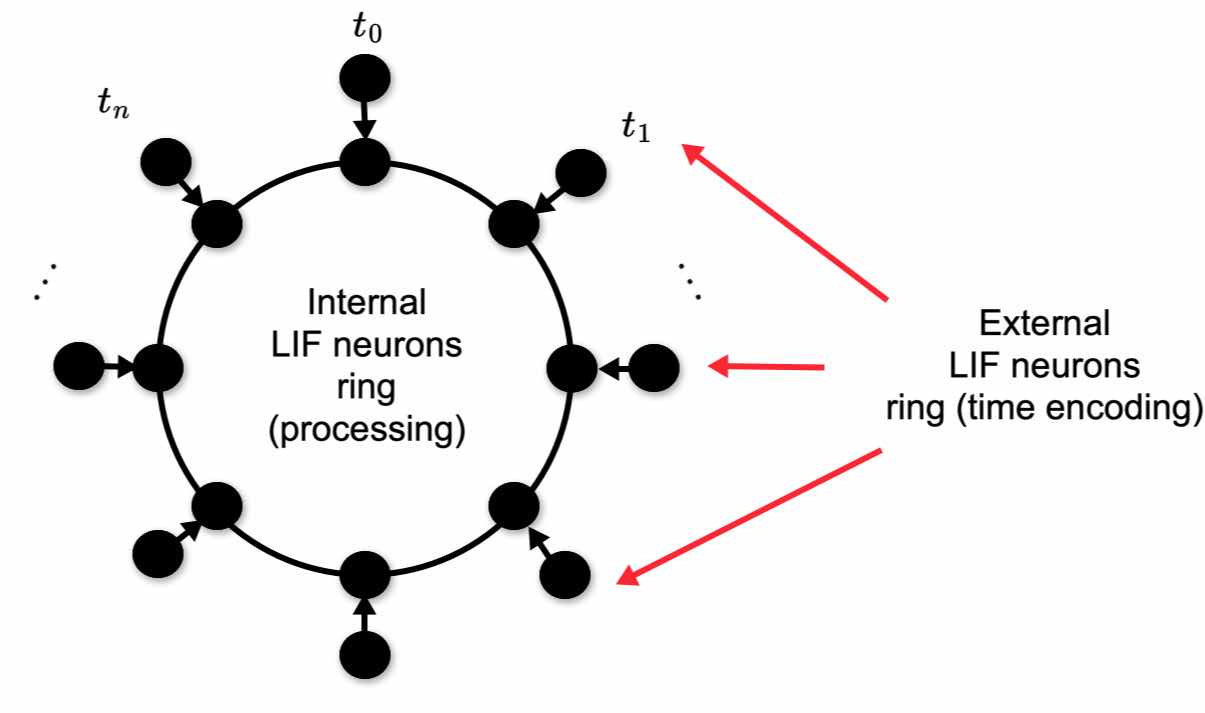}\includegraphics[scale = 0.36]{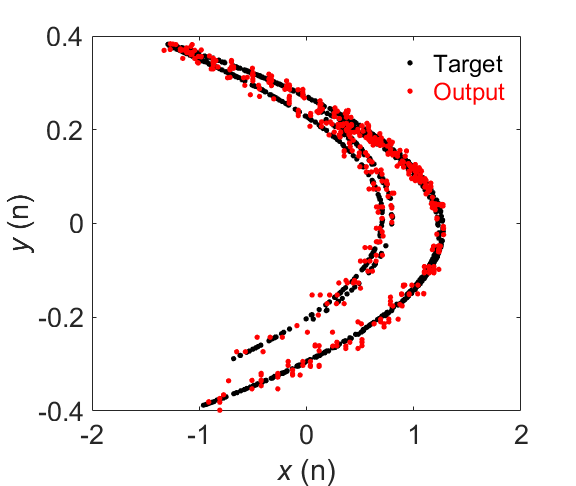}
    \caption{\textit{Left}: The hand-picked network, equivalent to a ring lattice with an additional external ring of input neurons.  \textit{Right}: Reconstructed attractor of the H\'enon map with the hand-picked network. This network consisted of $M = 100$ total neurons, $M_{in} = 50$ of them being the input neurons, and the remaining 50 making a ring connection with each neuron in the ring sending spikes to their immediate neighbors. The hand-picked network achieved $NRMSE = 0.0505$}
    \label{fig:ring}
\end{figure*}

After having introduced sparser and sparser random networks with non-satisfactory results even in the case of the $n+1$ time series, we have considered a hand-picked network. Looking back on the original equation of the H\'enon-map time series, we can see that the prediction of $u_{n+1}$ relies on knowledge of both $u_{n}$ and $u_{n-1}$. Therefore, the network must propagate spikes well after the initial $n$-th input has entered the network. To ensure that the memory of every input lingers in the network after the input has vanished, we propose the following network, shown in Fig. \ref{fig:ring} (left). The outer layer comprises $M_{in} = M/2$ neurons capable of receiving substantial currents from the input sequence. These outer neurons transmit spikes to the inner circle neurons, but no spikes are fed back to the outer ring. This design preserves the requirement for the outer ring to only retain information about the $n$ sequence of the time series.

Each neuron in the outer ring establishes a one-to-one connection with a neuron in the inner ring. The inner ring consists of $M/2$ neurons organized in a ring network. To ensure the spikes from $u_{n}$ continue to propagate while $u_{n+1}$ is applied, a delay of approximately $0.2*\delta$ is introduced for every spike. Consequently, when the network does the prediction of $u_{n}$, the inner ring contains information for $u_{n-1}$ sequence in the time series while the outer ring counts spikes of $u_{n}$. During the subsequent linear regression, the weights in $\textbf{W}_{out}$ that represent the $n$-th term and the $n-1$-th term are distinct so these linear weights can each specialize in approximating each term of the H\'enon map equation. With this arrangement, we saw that a spiking reservoir network could solve this task with a meaningful result. This network produced a normalized-root-mean-square-error of $NRMSE = 0.0505$.

This innovative network design enables efficient and distinct processing of temporal information, making it suitable for a variety of scientific applications.

\subsection*{Architecture meta-learning}
In order to validate that our proposed network was appropriate to solve time series tasks with spiking networks and reservoir computing, we used a meta-learning algorithm to see whether or not a random network would eventually approach our hand-picked network, or a similar network, and made insights about what the characteristics of an ideal network structure for time series tasks with reservoir computing spiking networks would be. We used the algorithm in Alg. \ref{alg:ml} below.

 \begin{algorithm}[H]
 \begin{algorithmic}[1]
 \renewcommand{\algorithmicrequire}{\textbf{Input: Temperature $T_0$, network $G$, steps $M$}}
 \renewcommand{\algorithmicensure}{\textbf{Output: network}}
 \REQUIRE in
 \ENSURE  out
 \\ \textit{Initialisation} :
  \STATE Given $G$, perform a reservoir computing task and obtain $RMSE=f_1$.
 \\ \textit{LOOP Process}
  \FOR {$i = M$ to $0$}
  \STATE Given $G$ Selected at random internal link and remove it, generate $G^\prime$. 
  \STATE Calculate the $RMSE=f_2$ with a reservoir computing learning task from $G^\prime$.
  \STATE Calculate $\Delta f=f_{new}-f_{old}$.
  \STATE Generate random number $r\in[0,1]$.
  \STATE Calculate $P=min(1,e^{-\Delta f/T(n)})$.
  \IF {($r \leq P$)}
  \STATE accept configuration, $G=G^\prime$
  \ENDIF
  \ENDFOR
 \RETURN $G$ 
 \end{algorithmic}
  \caption{Simulated annealing Meta-Learning}
 \label{alg:ml}
 \end{algorithm}


\subsection*{Meta-Learning for H\'enon map}
Following the meta-learning algorithm, we start with a network of $M = 100$ neurons, $50$ which are in the outer ring and $50$ which are in the inner network. The inner network started out as a random small world network with every neuron connected to its nearest neighbors $2$ units away. In addition, random connections are added to the inner network with a probability of $2/M$. The parameters for all the LIF neurons are the following: $V_{th} = 5$ and $\tau = 1$. The spiking parameters are the following: spike delay is $0.3 \delta$, the spike payload is $2$, $I_0 = 100$. The time step for the simulations is $\delta/200$ (NOTE: The parameters described above were used in all the H\'enon-Map simulations). $1000$ steps of this simulation were run with $T(n) = \frac{1}{50 n}$. 

\begin{figure}[htbp]
    \centering
    \includegraphics[width=0.49\linewidth]{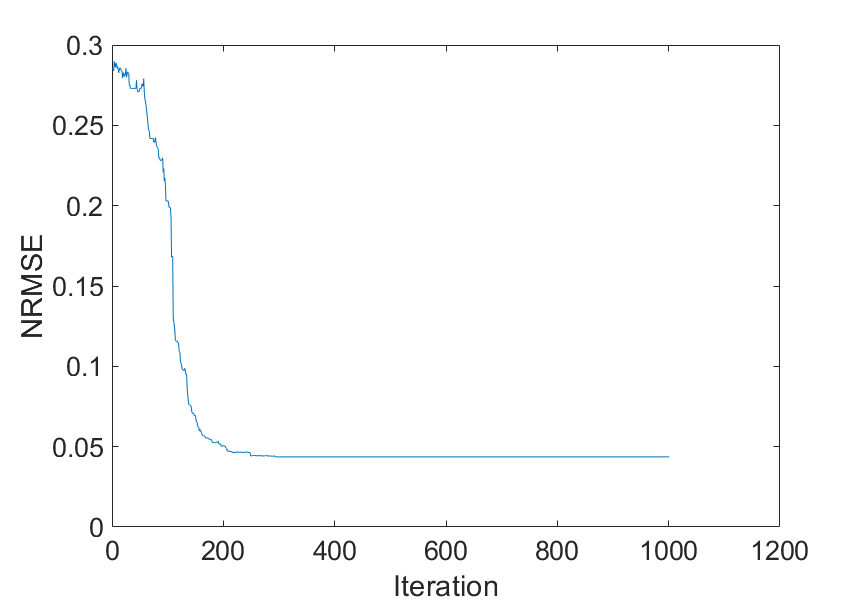}
    \includegraphics[width=0.49\linewidth]{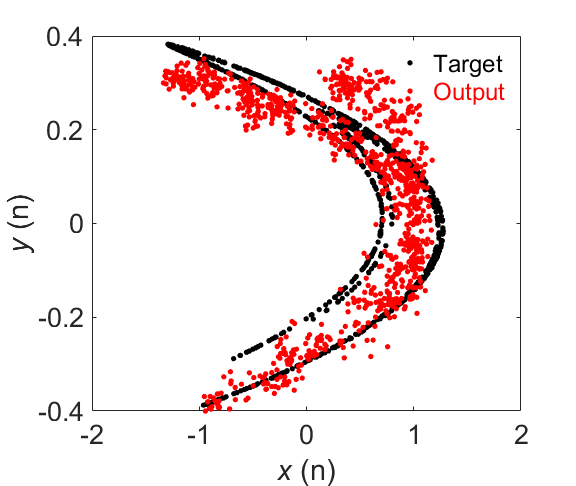}
    \includegraphics[width=0.49\linewidth]{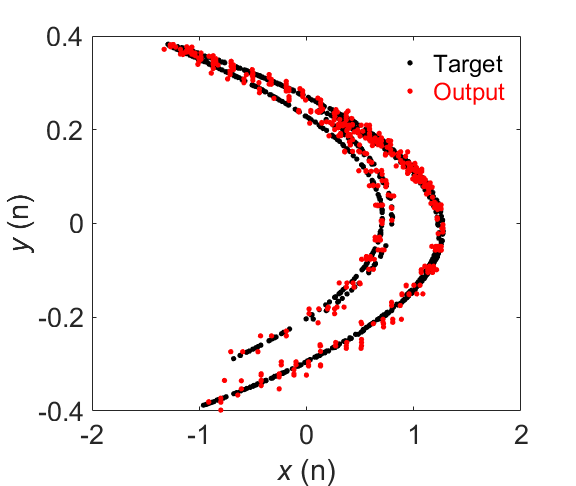}
    \includegraphics[width=0.49\linewidth]{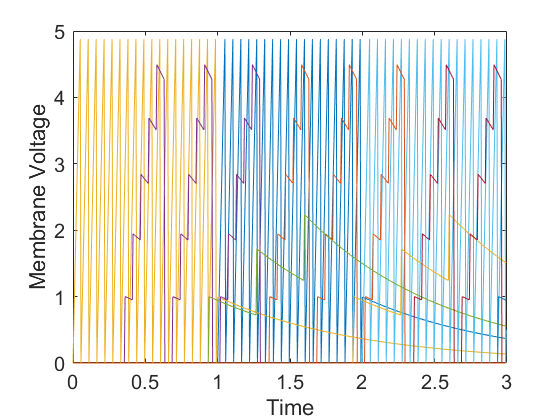}
    \caption{\textbf{Top left (TL).} NRMSE vs Iteration of meta-learning algorithm for the H\'enon Map Task. \textbf{Top right (TR) and Bottom Left (BL).} Two-dimensional representation of H\'enon Map prediction task results at the start of the meta-learning algorithm in \textbf{(TL)} and at the end of $1000$ steps in \textbf{TR.} \textbf{BR.} Membrane Voltage vs Time for each neuron in the network from time $[0,3]$.}
    \label{fig:meta-learning}
\end{figure}

\textbf{Results}. Meta-learning results of 1000 iterations are shown in \ref{fig:meta-learning}. In the final network, there were no longer any connections between neurons in the inner network, only connections between neurons in the outer ring and their one-to-one connections to the inner network. The connections that remained were only the one-to-one connections between the outer ring and the inner ring, forming chains of 2 neurons. The final $NRMSE$ of the network was $0.0437$.
This final network resembles aspects of our hand-picked network. In the hand-picked network, we proposed that the input neurons should have one-to-one connections with the rest of the network, and the rest of the network should have a simple connection scheme. The final network of the meta-learning also had these features, and to be specific, the internal network ended up being even simpler than a ring network. We can see that meta-learning has hinted that for time series, it is important for the network to have chains of neurons for signals to propagate far into the future. In the case of H\'enon-map, these chains ended up only being 2 neurons long, however, as we will show later in the paper in the case of the Mackey-Glass dataset, if the task requires deeper memory of past events to predict the future, this can be done simply by extending these chains to include more neurons.

\section*{Case of Mackey-Glass time series}

\textbf{Proposed Network for Mackey-Glass}
We now focus on the more complicated Mackey-Glass time series. Originally, we used the meta-learning algorithm above to see if the network could iteratively converge to the optimal network. However, this was shown to be unsuccessful. Either the algorithm was converging too slowly, or the appropriate network is unlikely to emerge from this algorithm.

\begin{figure}
    \centering
   \includegraphics[width=0.49\linewidth]{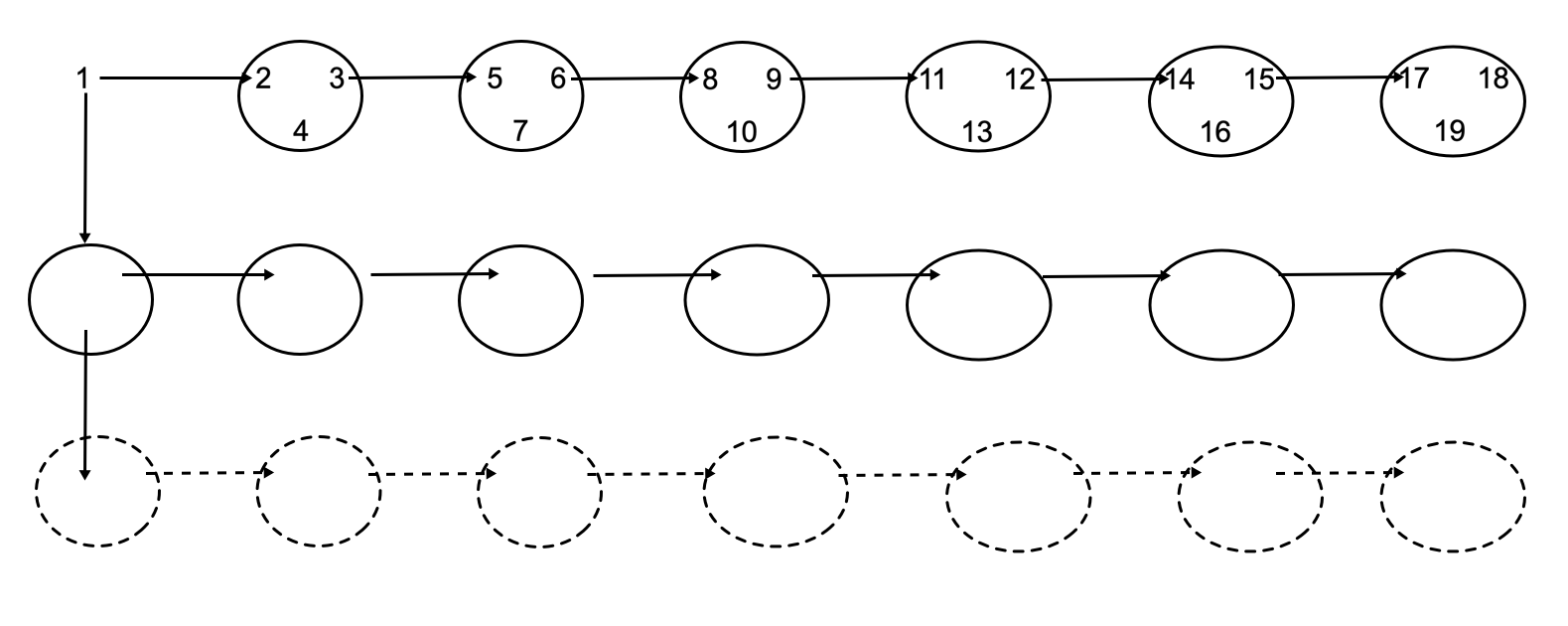}
   \includegraphics[width=0.49\linewidth]{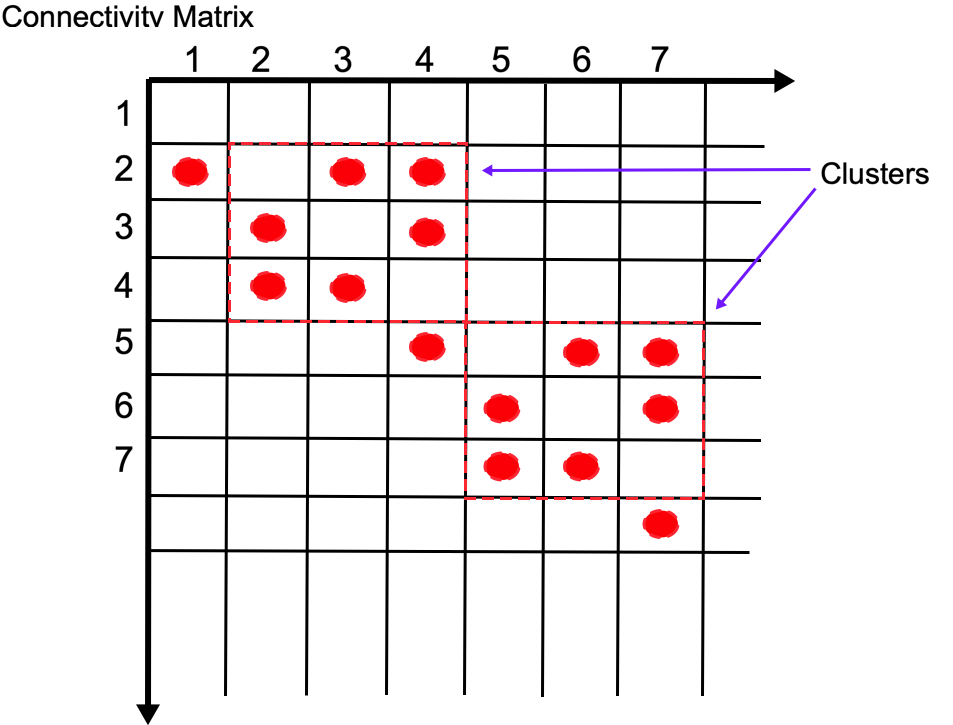}
    \caption{\textit{Left:} Proposed Network to solve Mackey-Glass Oscillator task. Each circle represents a cluster of neurons. \textit{Right:} Reservoir connection matrix for the proposed Mackey-Glass network. The $j^{th}$ row of the $i^{th}$ column represents the connection between the $j^{th}$ spike receiver and the $i^{th}$ spike sender. Non-zero values represent a connection, while values of $0$ represent no connection. In the illustration, $x$ represents connections, and empty values are zeros. The bolded 3x3 boxes represent each cluster of three. The pattern repeats along the diagonal for however many clusters are desired. Then the whole matrix is repeated along the diagonal for each input neuron desired.}
    \label{fig:connect1}
\end{figure}

Taking insight from our investigations of the H\'enon map task, we realized that the network for spiking reservoir networks for time series tasks must have long chains of neurons for the system to retain the memory of past inputs. Therefore we proposed the following network, shown in (figure \ref{fig:connect1}). This network has chains of clusters of neurons. For each input neuron, there is a chain of 6 clusters. In each cluster, there are 3 neurons. The idea behind this proposed network is that each chain facilitates memory up to $n-6$ in the past. In addition, the neurons that facilitate a memory up to $n-6$ in the past need to have a rich nonlinear response, employing multiple neurons, so that the first term in the Mackey-Glass Oscillator can be approximated well with a linear combination of their outputs. This proposed network does not necessarily accomplish this latter criterion sufficiently-- there is likely a better network that can be proposed which better achieves can approximate the first term of the original differential equation. In this example, we used 25 chains, which totaled a network of 475 neurons. The neuron and spike parameters are the following: $ V_{th} = 3, \tau = 0.5$ (time decay for all neurons), $\delta = 1$, $I_0 = 100$, the time delay for spikes is 2, and the spike payload is $2$.

\textbf{Results}.This network achieved a $NRMSE$ of $0.098$. This proposed model was able to solve the tasks decently well. When performing the same task with completely random networks, for a wide range of probabilities, this proposed network outperformed completely random networks which could only achieve a score of $0.155$ (although we only sampled 6 random graphs).


\section*{Loihi 2 on-chip implementation}

\textbf{Summary of Using Lava and Mackey-Glass Task}
A spiking neural network reservoir computing scheme was realized with the use of the Intel Lava python package \cite{intel-labs-lava-2023}.
Lava is an open-source software framework designed for creating neuro-inspired applications and aligning them with neuromorphic hardware. It equips developers with the necessary tools and abstractions to craft applications that leverage the fundamental principles of neural computation. To compile and execute processes across various backends, Lava leverages a low-level interface known as Magma, complemented by a robust compiler and runtime library provided by Intel, specifically tailored for use with the Loihi architecture. In addition, we were able to run our network on hardware using Intel virtual machines which have access to Loihi2 chips and NeuroCores.

\textbf{Lava Leaky Integrate and Fire Model}

Lava uses a discretized version of leaky integrate and fire (LIF) dynamics for its neurons. The LIF neurons behave with the following coupled equations:

\begin{lstlisting}[language=Python]
u[t] = u[t-1] * (1-du) + a_in        
v[t] = v[t-1] * (1-dv) + u[t] + bias 
s_out = v[t] > vth                    
v[t] = 0    
\end{lstlisting}
where u[t] is the applied current, v[t] is the membrane potential, du is the inverse of decay time-constant for current decay, dv is the inverse of decay time-constant for voltage decay, vth is the neuron threshold voltage, and bias is the constant bias applied to v[t].

Since the Loihi2 chip stores neuron information of state variables and parameters in the form of integers, the values assigned to parameters differ from the values they represent in the equations above. For example, when using Loihi2 Hardware configuration or the bit-accurate simulation of the Loihi2 chip, du, and dv can be assigned integer values between 0 and 4095, which correspond to values between 0 and 1, assigned values of vth correspond to multiples of 64, and integer spike payloads (1,2,3,...) correspond to an increase in v[t] in multiples of 64.

\begin{figure*}
    \centering
    \includegraphics[width=0.45\linewidth]{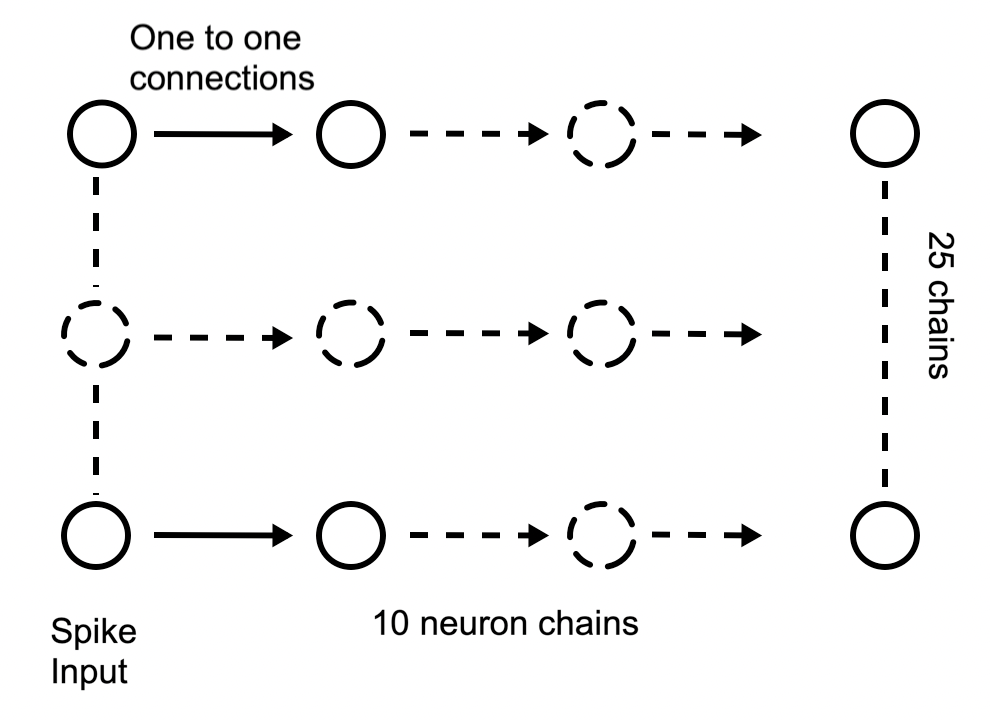}
    \includegraphics[width=0.45\linewidth]{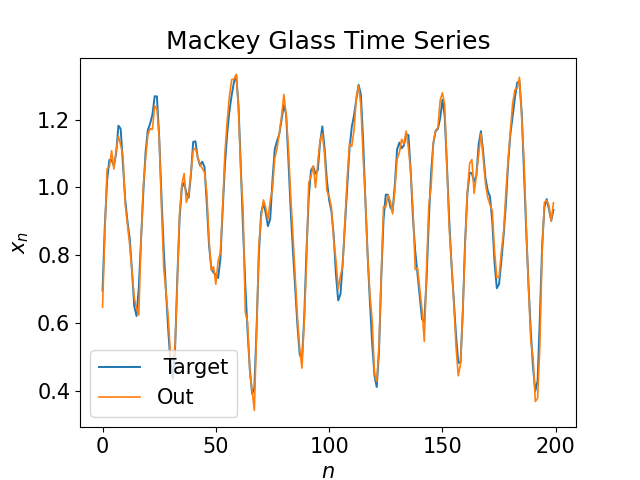}
    \caption{\textit{Left}: Proposed network for Mackey-Glass task. The network consists of an input layer connected to a reservoir made up of chains of 10 neurons. \textit{Right}: Performance of proposed network with Loihi2 on Mackey-Glass task. This network achieved a $NRMSE$ of $0.0338$. 
    }
    \label{fig:mackey-glass}
\end{figure*}

\textbf{Proposed Network}

The above shows the proposed network used to solve the Mackey-Glass Oscillator with the Lava package and Loihi2 NeuroChip. The input sequence $u_n$ is spatially encoded in the same way as before. In this case, there are $M_{in} = 25$ input neurons. Each of the $25$ input neurons is the first neuron of a ten-neuron chain. This network is a generalization of the network that was seen to be optimal for solving the H\'enon Map task. In the case of the H\'enon Map task, the chains were only 2 neurons long. Because the Mackey-Glass Oscillator requires longer memory, $6$, and time steps in the past, these chains were made longer. The output spikes are recorded using two additional layers of output neurons, which have one-to-one connections with the input layer and the reservoir, respectively. For each $u_n$, the input is applied to the network for $\delta$ discrete time steps and the membrane potential of the output layers is read out and reset for the next duration $\delta$.

Delays between spike sender and spike receiver were realized by utilizing a long current decay constant, $1/du$, which allows spikes to propagate up to $n+6$ after the initial spikes were applied to the reservoir.

$\textbf{More detail about Lava Parameters}$
We achieved a $NRMSE$ of $0.0338$ with our proposed network using Lava and Loihi2 hardware. The parameters are shown in Tab. \ref{tab:lava}.
\begin{table}[ht]
\centering
\begin{tabular}{|c|c|c|}
\hline
\texttt{Input\_LIF} & \texttt{Reservoir\_LIF} & \texttt{Output\_LIF} \\
\hline
shape = 25, & shape = 250, & shape = 25, and 250,\\
\texttt{bias\_mant  = 4} & \texttt{bias\_mant  = 0} & \texttt{bias\_mant  = 0}\\
vth = 1 (64) & vth = 82 (5248) & vth = 1000 (64000) \\
du = 4095 (1) & du = 80 (0.020) & du = 4095 (1) \\
dv = 0 & dv = 40 (0.0098) & dv = 0 \\
payload = 8 & payload = 8 &  \\
\hline
\end{tabular}
\caption{Parameters assigned to each LIF layer for Mackey Glass time series tasks using Lava and Loihi2. In parentheses are the actual values which they correspond to. Payload refers to the magnitude of the weights that connect spikes between the LIF layer to the \texttt{Reservoir\_LIF} layer.}
\label{tab:lava}
\end{table}


In this manuscript, we examined the implementation of reservoir computers using leaky integrate-and-fire neurons, particularly on the Loihi chip architecture. The focus has been on the H\'enon and Mackey-Glass prediction tasks, emphasizing the critical role of designed networks in achieving high performance. The study began with a deliberate but thought selection of network architectures, starting from a stochastic dense network ( Erd\H{o}s-R\'enyi), then transitioning to a small-world networks-like (inspired by Watts-Strogatz model), and progressively refining towards even sparser models to enhance performance.  

We have also considered a meta-learning framework that was developed, with the network selection determined through simulated annealing of the model's fitness (NRMSE minimization), using a fixed task. Through meta-learning, we have demonstrated a streamlined process for finding (quasi)optimal architectures for tasks like fitting the H\'enon map and predicting Mackey-Glass time series. This not only conserves valuable human resources but also results in architectures finely tuned to the underlying data distributions. Meta-learning has nonetheless proven (in this case, and with this implementation) to be less powerful than our hand-designed networks. The findings of this study highlight the efficacy of manual network construction in recurrent neural network design, although within the set of limited tasks we considered. As a drawback,  such a traditional manual network design approach can be time-consuming.  

While this study successfully hints towards the benefits of manual network construction, in conjunction with Intel Lava and Loihi, there are promising directions for future research. For instance, scaling these approaches to more complex tasks (such as classification) and larger datasets could provide insights into their broader applicability.

\subsection*{Energy consumption with Loihi2}

Here, we investigate the energy consumption of the reservoir computing model implemented on the Intel Loihi 2 neuromorphic chip for the optimal reservoir and use of parameters in Tab. \ref{tab:lava}. The results were not noticeably different from the results of Fig. \ref{fig:mackey-glass}. We then conducted energy measurements directly on the Loihi 2 chip, following the structure and parameters outlined in our original simulations using the Lava architecture and for the same optimized reservoir.

The energy measurements were taken using Lava's built-in \textit{Profiler} module in Intel's Loihi 2 Oheogulch board. The reservoir model occupies 6 neuromorphic cores and 1 Loihi 2 chip using Lava on Loihi version 0.5.0. These energy measurements provide the average energy and power used by the algorithm. Measurements were taken per algorithmic step maintaining the original goal of $1000$ epochs and an interval of $90$ Loihi timesteps, as specified in the initial implementation. The energy consumption data was collected from 1 to 1000 epochs in increments of approximately 33 epochs. This approach was chosen due to the impracticality of measuring energy consumption for every single epoch within the range, given the time constraints posed by the physical chip.  

The setup for these measurements is detailed as follows:
\begin{itemize}
    \item \textbf{Number of epochs:} 1000
    \item \textbf{Interval:} 90 Loihi timesteps
    \item \textbf{Measurement increments:} ~33 epochs
\end{itemize}

\begin{figure*}
    \centering
    \includegraphics[width=0.99\linewidth]{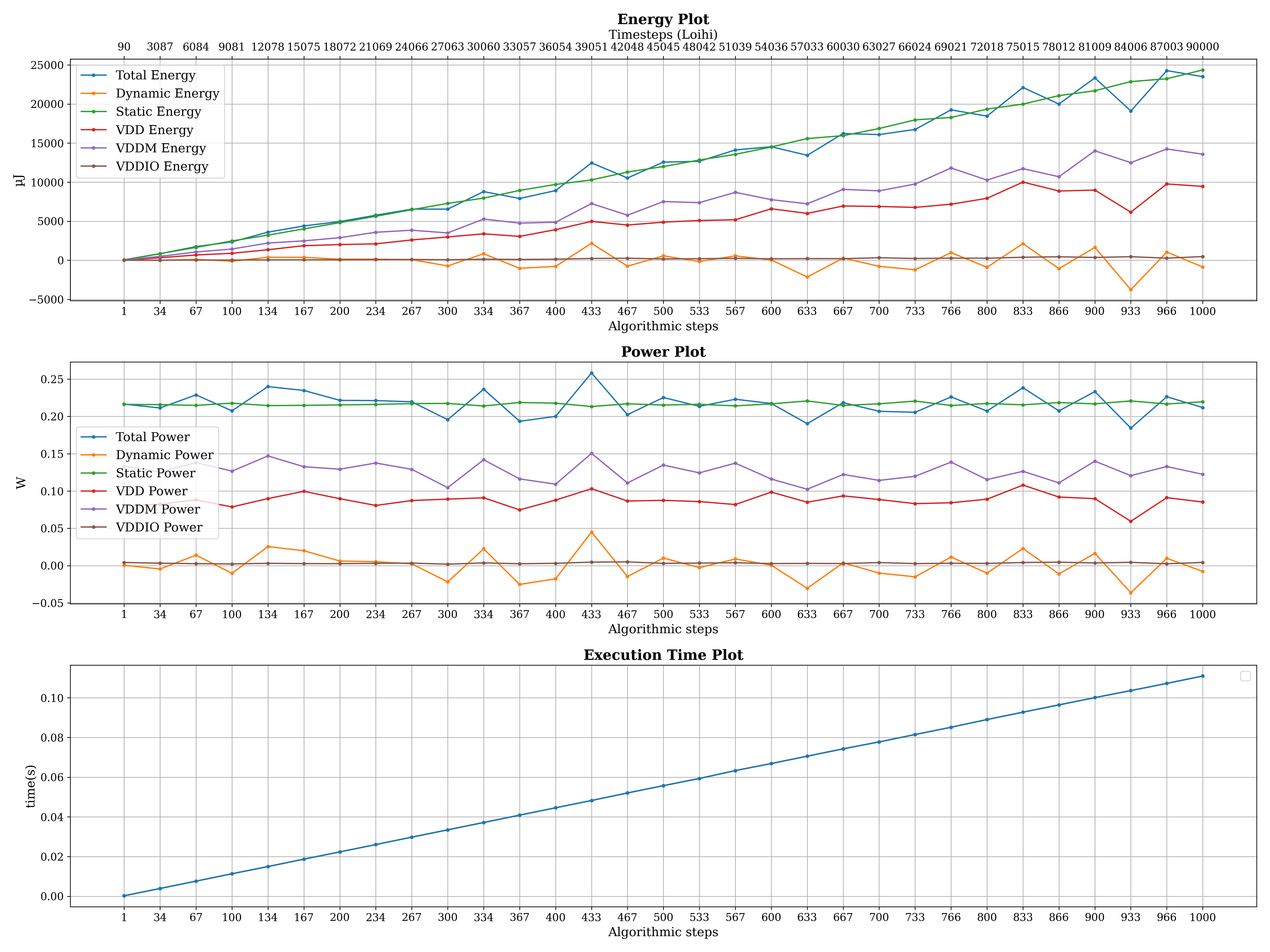}
    \caption{Energy consumption of the reservoir computing model implemented on the Intel Loihi neuromorphic chip. Measurements were taken from 1 to 1000 epochs in increments of approximately 33 epochs, with a delta of 90 Loihi timesteps per epoch. The plot illustrates the energy usage trends throughout the training process.}

    \label{fig:energy_measurements}
\end{figure*}

Results of this experiment are detailed in Fig. \ref{fig:energy_measurements}, where we depict the energy consumption pattern as training progresses through the epochs. A summary is provided in 
Tab. \ref{tab:energyc}. 

Tab. \ref{tab:energyPerEpoch} shows the energy consumption per epoch, where the energy per time step is the total energy consumed per timestep in Loihi. A Loihi timestep is the time between spikes. The throughput is the measure of how many Loihi timesteps can be processed per second. The table also details the cumulative energy consumed at a specific epoch for the whole training process.

Total energy is the sum of the dynamic and static energy. Static energy refers to the transistor leakage energy or the energy consumed when the system is idle. Dynamic energy is the energy consumed by active operations of transistors during component switching. The VDD energy measurements correspond to the built-in Lakemont x86 in Loihi. The VDDM energy is the mean energy consumed by SRAM over time, VDDIO is the mean energy consumed by the FPIO/IO interfaces, and the VDD energy is the mean energy consumed by the logic power.

\begin{table}[h]
    \centering
    \caption{Performance Benchmarking Comparison Template}
    \label{tab:benchmark}
    \begin{tabular}{>{\centering\arraybackslash}p{2cm} 
                    >{\centering\arraybackslash}p{2cm} 
                    >{\centering\arraybackslash}p{2cm} 
                    >{\centering\arraybackslash}p{2cm} 
                    >{\centering\arraybackslash}p{2cm} 
                    >{\centering\arraybackslash}p{2cm}}
        \toprule
        Epoch & Energy per timestep ($\mu$J/operation) & Throughput (timestep/s) & \multicolumn{3}{c}{Energy per epoch ($\mu$J)} \\
          \addlinespace[-3.5ex]
        \cmidrule{4-6}
        & & & Static & Dynamic & Total \\ 
        \midrule
        1 & 0.269 & 784099 & 832 & 18 & 850 \\
        500 & 0.278 &  808243 & 12558 & 563 & 11994 \\
        1000 & 0.261 & 811205 & 23502 & 859 & 24362  \\
        \bottomrule
    \end{tabular}
    \label{tab:energyPerEpoch}
\end{table}

\begin{table}[h!]
\centering
\caption{Summary of Energy, Power, and Execution Time Measurements. The energies are estimated both per 34 timesteps and in W.}
\begin{tabular}{|l|c|c|}
\hline
\textbf{Metric} & \textbf{Approximate Value} & \textbf{Unit} \\ \hline
Total Energy Increase & 25000 & $\mu$J/34ts \\ \hline
Dynamic Energy Increase & 5000 & $\mu$J/34ts \\ \hline
Static Energy Increase & 15000 & $\mu$J/34ts\\ \hline
VDD Energy Increase & 10000 & $\mu$J//34ts\\ \hline
VDDM Energy Increase & 5000 & $\mu$J/34ts \\ \hline
VDDIO Energy Increase & 2500 & $\mu$J/34ts \\ \hline
Total Power & 0.25 & W \\ \hline
Dynamic Power & 0.05 & W \\ \hline
Static Power & 0.15 & W \\ \hline
VDD Power & 0.10 & W \\ \hline
VDDM Power & 0.05 & W \\ \hline
VDDIO Power & 0.025 & W \\ \hline
Execution Time per Timestep & 0.001 & seconds \\ \hline
\end{tabular}
\label{tab:energyc}
\end{table}

\begin{figure}
    \centering
    \includegraphics[width=0.95\linewidth]{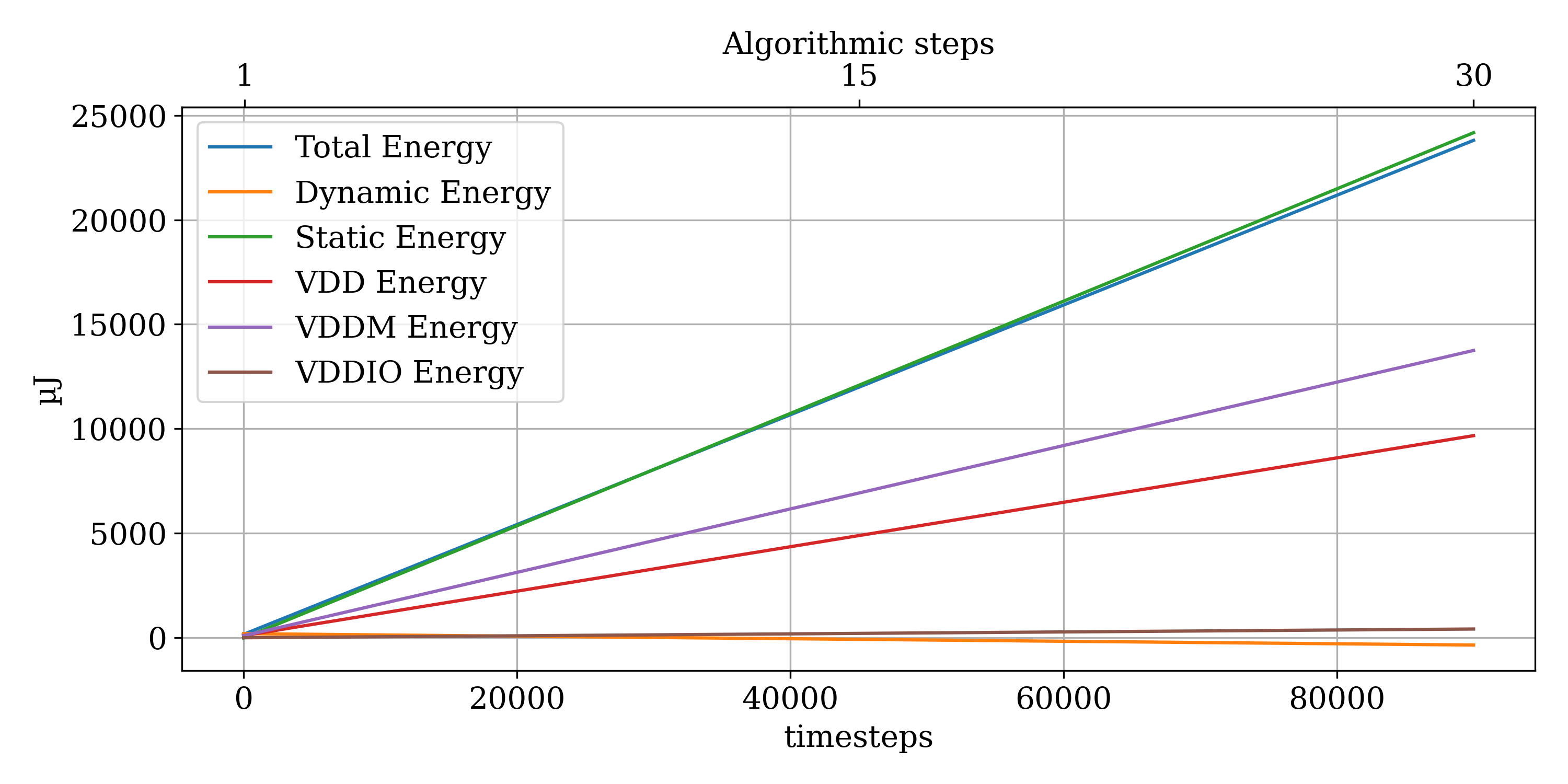}
        \includegraphics[width=0.95\linewidth]{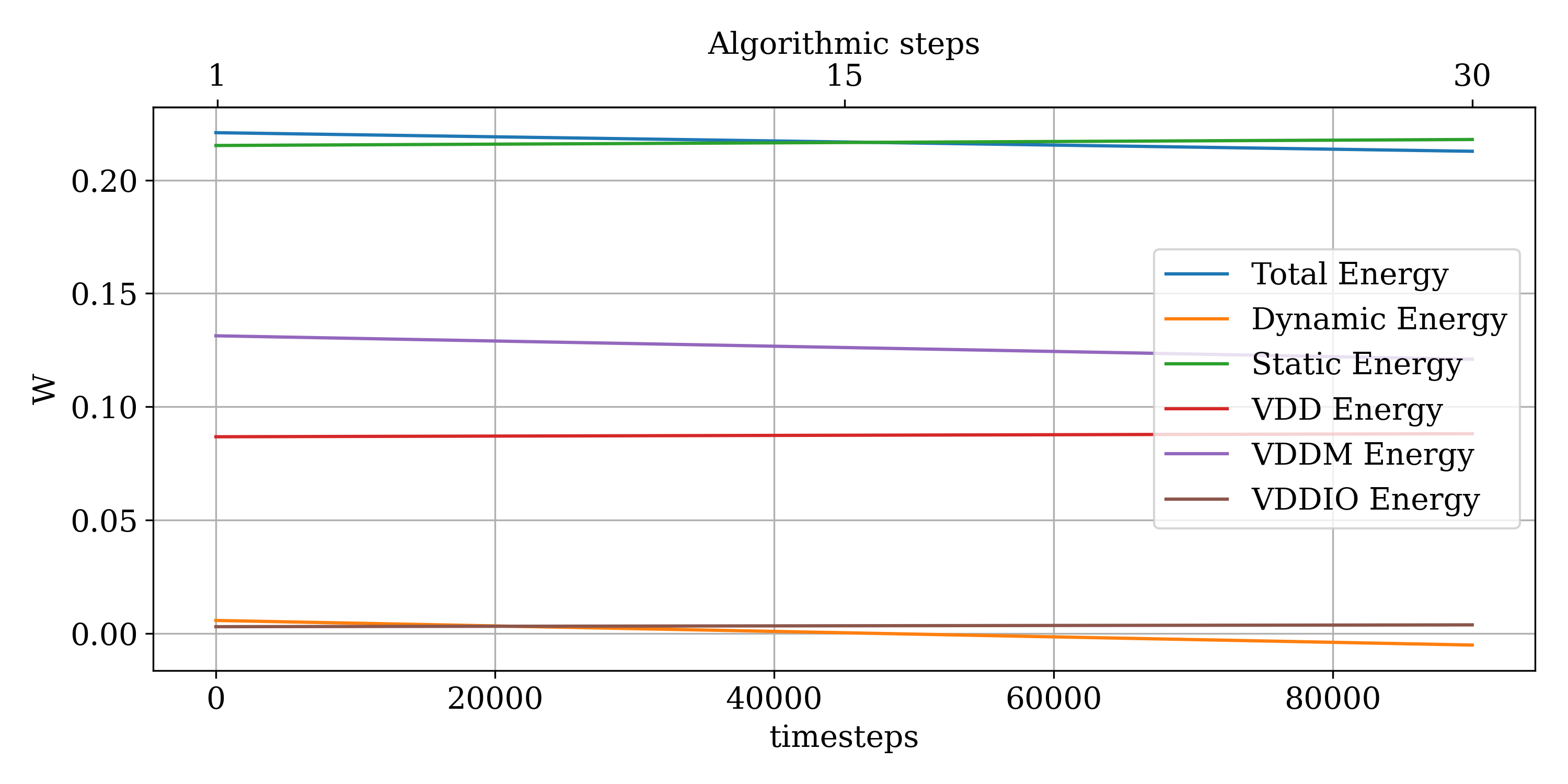}
    \caption{Estimated energy and power as a function of the number of timesteps, obtained from Tab. \ref{tab:energyc} or Fig. \ref{fig:energy_measurements} from a single cycle.}
    \label{fig:enregression}
\end{figure}

Using Fig.  \ref{fig:energy_measurements} we
can estimate the energy consumption throughout the reservoir execution via a regression.  We provide these results in Fig. \ref{fig:enregression}. 

As we can see, implementing the reservoir computer has an energy expenditure of roughly $0.25$ W in terms of total power, running at roughly $1$ms per timestep. This is roughly in the bulk part of other machine learning algorithms run on Loihi, for instance, neural backpropagation \cite{sornborger}, although for a completely different task.

The results give a glimpse of the potential applications and advantages of SNNs in terms of energy efficiency, in particular, compared to the hundreds of W of a current generation of GPU.

\section*{Conclusions}
To conclude, our results can be summarized as follows. Sparser networks of LIF neurons perform better when it comes to the prediction task of both the H\'enon and the Mackey-Glass time series, using a reservoir computing architecture. As we have shown, RC can be directly implemented on the Loihi architecture, which in this work has provided the best NMRSE for the Mackey-Glass time series prediction, although with an optimized network architecture. For completeness, the NRMSE for each network is shown in Table. \ref{table:final}.

\begin{table}
\centering
\begin{tabular}{c|c|c}
Time Series & Network & NRMSE \\
\hline
   H\'enon map  &  Random architecture (best) & 0.17 \\
   H\'enon map  &  Hand picked architecture - Fig. \ref{fig:ring} & 0.05 \\
     H\'enon map  & Meta-Learning  - Fig. \ref{fig:meta-learning}& 0.05 \\
  Mackey-Glass & Random architecture (best) & 0.15 \\
     Mackey-Glass & Hand picked architecture - Fig \ref{fig:connect1} & 0.09 \\
     Mackey-Glass & Hand picked architecture (Loihi2)  - Fig \ref{fig:mackey-glass} & 0.03
\end{tabular}
\caption{Summary of the prediction task NMRSE for different architectures and two time series.  }
\label{table:final}
\end{table}

Exploring the synergy of meta-learning with other optimization techniques and incorporating domain-specific constraints will be the focus of future works.

\section*{Acknowledgments}
The work of SK, AS, and FC was carried out under the auspices of the NNSA of the U.S. DoE at LANL under Contract No. DE-AC52-06NA25396. FC acknowledges support from LDRD via 20230338ER and 20230627ER.

\bibliography{library}

\bibliographystyle{abbrv}

\end{document}